\documentclass[runningheads]{llncs}
\usepackage{graphicx}
\usepackage{eucal}
\usepackage{booktabs}
\usepackage{todonotes}
\usepackage{verbatim}
\usepackage{amsmath}
\usepackage{amssymb}
\usepackage[normalem]{ulem}
\useunder{\uline}{\ul}{}

\usepackage{todonotes}
\newcommand{\mathilde}[1]{\todo[fancyline,backgroundcolor=green!25,bordercolor=green]{mathilde says: #1}}
\newcommand{\herve}[1]{\todo[fancyline,backgroundcolor=red!25,bordercolor=red]{Herv\'{e} says: #1}}

\begin{document}

\title{Source-Relaxed Domain Adaptation for~Image~Segmentation}


\author{Mathilde Bateson \and Hoel Kervadec \and Jose Dolz \and  Herv\'e Lombaert \and Ismail Ben Ayed}


%

\institute{ETS Montr\'eal\\ \email{mathilde.bateson.1@ens.etsmtl.ca}}

\authorrunning{M. Bateson et al.}


\maketitle


\newcommand{\mathbbm}[1]{\text{\usefont{U}{bbm}{m}{n}#1}}



\begin{abstract}

Domain adaptation (DA) has drawn high interests for its capacity to adapt a model trained on labeled source data to perform well on unlabeled or weakly labeled target data from a different domain. Most common DA techniques require the concurrent access to the input images of both the source and target domains. However, in practice, it is common that the source images are not available in the adaptation phase. This is a very frequent DA scenario in medical imaging, for instance, when the source and target images come from different clinical sites. We propose a novel formulation for adapting segmentation networks, which relaxes such a constraint. Our formulation is based on minimizing a label-free entropy loss defined over target-domain data, which we further guide with a domain-invariant prior on the segmentation regions. Many priors can be used, derived from anatomical information. Here, a class-ratio prior is learned via an auxiliary network and integrated in the form of a Kullback–Leibler (KL) divergence in our overall loss function. We show the effectiveness of our prior-aware entropy minimization in adapting spine segmentation across different MRI modalities. Our method yields comparable results to several state-of-the-art adaptation techniques, even though is has access to less information, the source images being absent in the adaptation phase. Our straight-forward adaptation strategy only uses one network, contrary to popular adversarial techniques, which cannot perform without the presence of the source images. Our framework can be readily used with various priors and segmentation problems.

\end{abstract}

\section{Introduction}

Semantic segmentation, or the pixel-wise annotation of an image, is a key first step in many clinical applications. Since the introduction of deep learning methods, automated methods for segmentation have outstandingly improved in many natural and medical imaging problems \cite{litjens2017survey}. Nonetheless, the pixel-level ground-truth labelling necessary to train these networks is time-consuming, and deep-learning methods tend to under-perform when trained on a dataset with an underlying distribution different from the target images. To circumvent those impediments, methods learning robust networks with less supervision have been popularized in computer vision.

\begin{figure}[t]
    \includegraphics[width=1\linewidth]{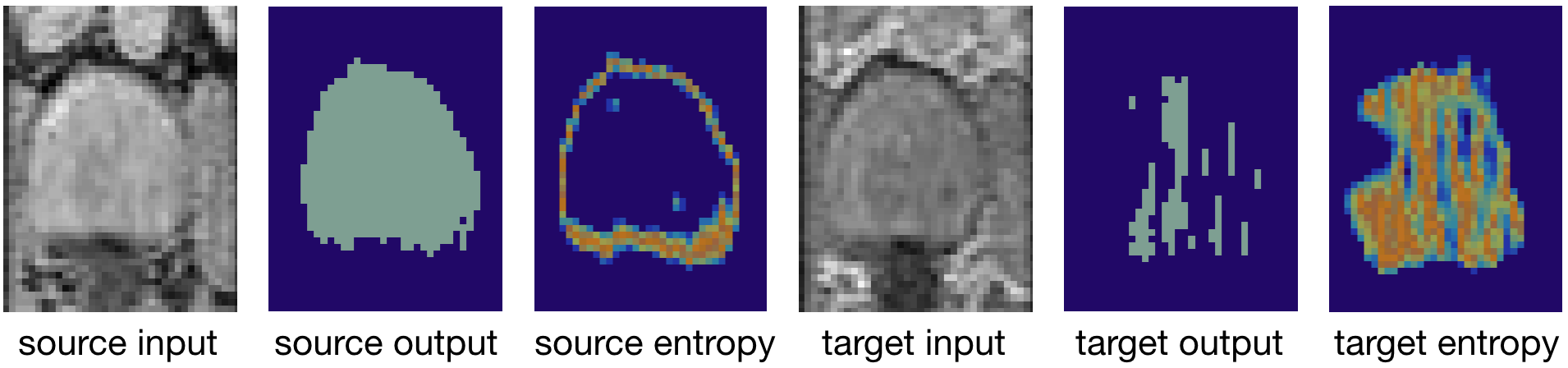}
      \caption[]{Visualization of 2 aligned slice pairs in source (Water) and target modality (In-Phase): the domain shift in the target produces a drop in confidence and accuracy.}
         \label{fig:s_t_im}
\end{figure}

Domain Adaptation (DA) adresses the transferability of a model trained on an annotated source domain to another target domain with no or minimal annotations. The presence of a domain shift, such as those produced by different protocols, vendors, machines in medical imagining, often leads to a big performance drop (see Fig. \ref{fig:s_t_im}). Adversarial strategies are currently the prevailing techniques to adapt networks to a different target domain, both in natural \cite{gholami2018novel,javanmardi2018domain,kamnitsas2017unsupervised,zhao2019supervised} and medical \cite{chen2018road,hoffman2017cycada,hong2018conditional,tsai2018learning} images. These methods can either be generative, by transforming images from one domain to the other \cite{CycleGAN2017}, or can minimize the discrepancy in the feature or output spaces learnt by the model \cite{DouPnP,tzeng2017adversarial,tsai2018learning}. 

One major limitation of adversarial techniques is that, by design, they require the concurrent access to both the source and target data during the adaptation phase. 
In medical imaging, this may not be always feasible when the source and target data come from different clinical sites, due to, for instance, privacy concerns or the loss or corruption of source data. Amongst alternative approaches to adversarial techniques, self training \cite{zou2018unsupervised} and the closely-related entropy minimization \cite{advent,Wu2020EntropyMV,morerio2018minimalentropy} were investigated in computer vision. As confirmed by the low entropy prediction maps in Fig. \ref{fig:s_t_im}, a model trained on an imaging modality tends to produce very confident predictions on within-sample examples, whereas uncertainty remains high on unseen modalities. 
As a result, enforcing high confidence in the target domain as well can close the performance gap. This is the underlying motivation for entropy minimization, which was first introduced in semi-supervised \cite{Grandvalet} and unsupervised \cite{GomesNIPS2010} learning. 
To prevent the well-known collapse of entropy minimization to a trivial solution with a single class, the recent domain-adaptation methods in \cite{advent,Wu2020EntropyMV} further incorporate a criterion encouraging diversity in the prediction distributions. However, similarly to adversarial approaches, the entropy-based methods in \cite{advent,Wu2020EntropyMV} require access to the source data (both the images and labels) during the adaptation phase via a standard supervised cross-entropy loss. The latter discourages the trivial solution of minimizing the entropy alone on the unlabeled target images.

We propose a domain-adaptation formulation tailored to a setting where the source data is unavailable (neither images, nor labeled masks) during the training of the adaptation phase. Instead, our method only requires the parameters of a model previously trained on the source data for initialisation. 
Our formulation is based on minimizing an label-free entropy loss defined over target-domain data, which we  further guide with a domain-invariant prior on the segmentation regions. Many priors can be used, derived from  anatomical information. Here, a class-ratio prior is learned via an auxiliary network and integrated in the form of a Kullback–Leibler (KL) divergence in our overall loss function. Unlike the recent entropy-based methods in \cite{advent,Wu2020EntropyMV}, our overall loss function relaxes the need to access to the source images and labels during adaptation, as we do not use a source-based cross-entropy loss. Our class-ratio prior is related to several recent works in the context of semi- and weakly-supervised learning \cite{yuille,kervadec2019constrained}, which showed the potential of domain-knowledge priors for guiding deep networks when labeled data is scarce. Also, the recent works in \cite{zhang2019curriculum,Bateson2019} integrated priors on class-ratio/size in domain adaptation but, unlike our work, in the easier setting where one has access to source data (both the images and labels). In fact, the works in \cite{zhang2019curriculum,Bateson2019} used a cross-entropy loss over labeled source images during the training of the adaptation phase. 

We report comprehensive experiments and comparisons with state-of-the-art domain-adaptation methods, which show the effectiveness of our prior-aware entropy minimization in adapting spine segmentation across different MRI modalities. Surprisingly, even though our method does not have access to source data during adaptation, it achieves better performances than the state-of-the-art methods in \cite{zhang2019curriculum,tsai2018learning}, while greatly improving the confidence of network predictions. 
Our framework can be readily used for adapting any segmentation problems. Our code is publicly and anonymously available \footnote{https://github.com/mathilde-b/SRDA}. To the best of our knowledge, we are the first to investigate domain adaptation for segmentation without direct access to the source data during the adaptation phase.






\begin{figure}[t]
    \includegraphics[width=1\linewidth]{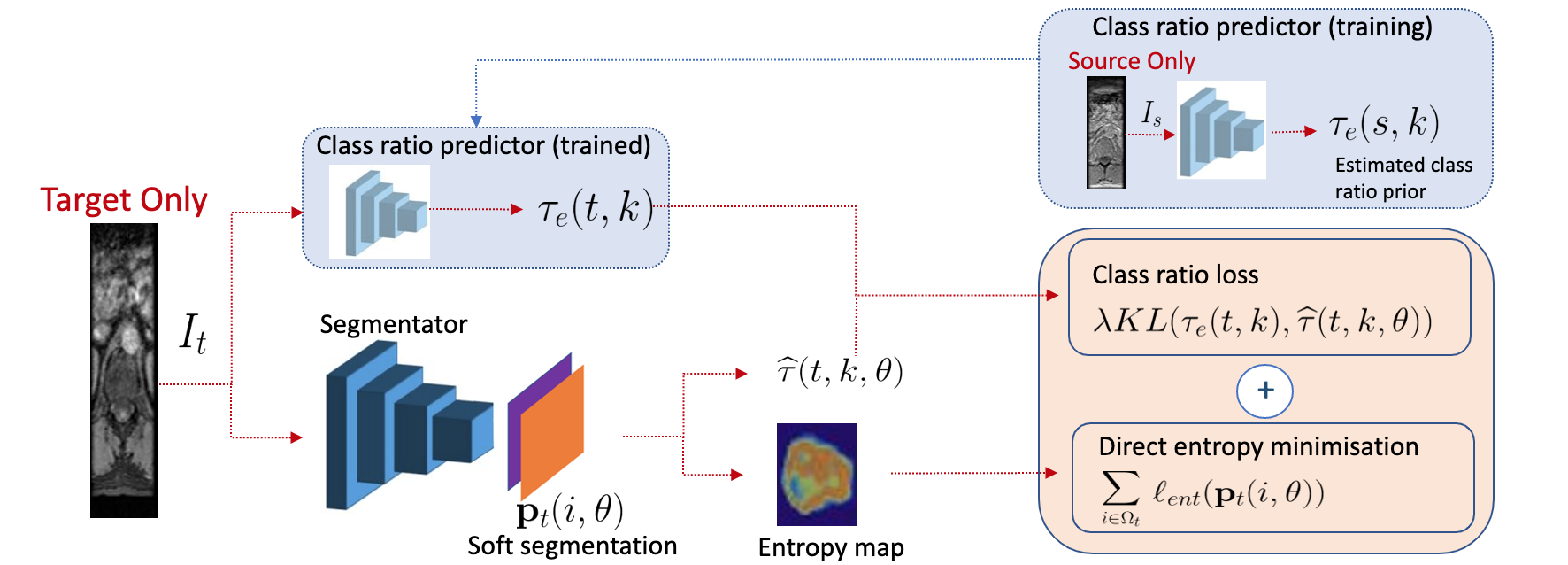}
      \caption[]{Overview of our framework for Source-Relaxed Domain Adaptation: we leverage entropy minimization and a class-ratio prior to relax the need for a concurrent access to the source and target data.}
      \label{fig:overview}
\end{figure}

\section{Method} 

We consider a set of $S$ source images ${I}_s: \Omega_s\subset \mathbb R^{d} \rightarrow {\mathbb R}$, $d\in \left \{ 2,3 \right \}$, $s=1, \dots, S$. The ground-truth K-class segmentation of $I_s$ can be written, for each pixel (or voxel) $i \in \Omega_s$, as a simplex vector ${\mathbf y}_s (i) = (y^1_s (i), \dots, y^K_s (i)) \in \{0,1\}^K$.
For domain adaptation (DA) problems, the network is usually first trained on the source domain only, by minimizing a standard supervised loss with respect to network parameters $\theta$:
\begin{equation}\label{eq:source-sup}
\begin{aligned}
\mathcal{L}_{s}\left(\theta, \Omega_{s}\right)= \frac{1}{\left|\Omega_{s}\right|} \sum_{s=1}^{S} \ell\left({\mathbf y}_s (i), {\mathbf p}_s (i, \theta)\right)
  \end{aligned}
\end{equation}
where ${\mathbf p}_s (i, \theta) = (p^1_s (i,\theta), \dots, p^K_s (i, \theta)) \in [0,1]^K$ is the softmax output of the network at pixel/voxel $i$ in image $I_s$, and $\ell$ is the standard cross-entropy loss: $\ell({\mathbf y}_s (i), {\mathbf p}_s (i, \theta)) = - \sum_k y^k_s (i) \log p^k_s (i, \theta)$.

Given $T$ images of the target domain, ${I}_t: \Omega_t\subset \mathbb R^{2,3} \rightarrow {\mathbb R}$, $t=1, \dots, T$,
the first loss term in our adaptation phase encourages high confidence in the softmax predictions 
of the target, which we denote ${\mathbf p}_t (i, \theta) = (p^1_t (i,\theta), \dots, p^K_t (i, \theta)) \in [0,1]^K$. This is done by minimizing the entropy of each of these predictions: 
\begin{equation}
\label{ent-target}
\ell_{ent}({\mathbf p}_t (i,\theta)) = - \sum_k p^k_t (i,\theta) \log p^k_t (i, \theta)
\end{equation}
However, it is well-known from the semi-supervised and unsupervised learning literature \cite{Grandvalet,GomesNIPS2010,jabi} that minimizing this entropy loss alone may result into degenerate trivial solutions, biasing the prediction towards a 
single dominant class. To avoid such trivial solutions, the recent domain-adaptation works in \cite{advent,Wu2020EntropyMV} integrated a standard supervised cross-entropy loss over the source data, i.e., Eq. \eqref{eq:source-sup}, during the training of the adaptation phase. However, this requires access to the source data (both the images and labels) during the adaptation phase. To relax this requirement, we embed domain-invariant prior knowledge to guide unsupervised entropy training during the adaptation phase, which takes the form of a class-ratio (i.e. region proportion) prior. We express the (unknown) true class-ratio prior for class $k$ and image $I_t$ as: $ \tau_{GT}(t,k) = \frac{1}{\left|\Omega_{t}\right|}\sum_{i \in \Omega_t}  y^k_t (i)$. As the ground-truth labels are unavailable in the target domain, this prior cannot be computed directly. Instead, we train an auxiliary network on source data to produce an estimation of class-ratio in the target\footnote{Note that many other estimators could be used, e.g., using region statistics from the source domain or anatomical prior knowledge.}, which we denote $\tau_{e}(t,k)$. 
Furthermore, the class-ratio can be approximated from the the segmentation network's output for target image $I_t$ as follows: $\widehat{\tau}(t,k,\theta)=\frac{1}{\left|\Omega_{t}\right|} \sum_{i \in \Omega_t} p^k_t (i, \theta)$

To match these two probabilities representing class-ratios, we integrate a KL divergence with the entropy in Eq. \eqref{ent-target}, minimizing the following overall loss during the training of the adaptation phase:
\begin{equation}\label{eq:ent-domain-adaption}
\begin{aligned}
  \min_{\theta}\sum_{t} \sum_{i \in \Omega_t} \ell_{ent}({\mathbf p}_t (i, \theta))+ \lambda \mbox{KL}(\tau_{e}(t,k),\widehat{\tau}(t,k,\theta))
  \end{aligned}
\end{equation}
Clearly, minimizing our overall loss in Eq. \eqref{eq:ent-domain-adaption} during adaptation does not use the source images and labels, unlike the recent entropy-based domain adaptation methods in \cite{advent,Wu2020EntropyMV}.


\section{Experiments}
    
\subsection{Cross-Modality Adaption with entropy minimization} 

\subsubsection{\textbf{Dataset.}} We evaluated the proposed method on the publicly available MICCAI 2018 IVDM3Seg Challenge\footnote{https://ivdm3seg.weebly.com/} dataset of 16 manually annotated 3D multi-modal magnetic resonance scans of the lower spine, in a study investigating intervertebral discs (IVD) degeneration. We first set the water modality (Wat) as the source and the in-phase (IP) modality as the target domain (Wat $\rightarrow$ IP), then reverted the adaptation direction (IP $\rightarrow$ Wat). 13 scans were used for training, and the remaining 3 scans were kept for validation. The slices were rotated in the transverse plane, and no other pre-processing was performed. The setting is binary segmentation (K=2), and the training is done with 2D slices.

\subsubsection{\textbf{Benchmark Methods}}

We compare our loss to the recent one adopted in \cite{zhang2019curriculum}:
\[\mathcal{L}_{s}\left(\theta, \Omega_{s}\right)+ \lambda \sum_{t} \sum_{i \in \Omega_T} KL(\tau_{e}(t,k),\widehat{\tau}_{t}(t,k,\theta))\]
Note that, in \cite{zhang2019curriculum}, the images from the source and target domain must be present concurrently in this framework, which we denote $AdaSource$. 
We also compared to the state-of-the art adversarial method in \cite{tsai2018learning}, denoted $Adversarial$. 
A model trained on the source only with Eq\eqref{eq:source-sup}, \textit{NoAdaptation}, was used as a lower bound. A model trained with the cross-entropy loss on the target domain, referred to as $Oracle$, served as an upper bound.

\subsubsection{\textbf{Learning and estimating the class-ratio prior.}}\label{sec:sizeprior}

We learned an estimation of the ground-truth class-ratio prior via an auxiliary regression network $R$, which is trained on the images $I_s$ from the source domain $S$, where the ground-truth class-ratio $\tau_{GT}(s,k)$ is known. $R$ is trained with the squared $\mathcal{L}_{2}$ loss: $\min _{\tilde{\theta}} \sum_{s=1...S}\left(R(I_s, \tilde{\theta})-\tau_{GT}(s,k)\right)^{2} \label{eq:l2}$. The estimated class-ratio prior $\tau_{e}(t,k)$ of an image $I_t$ in the target domain is obtained by inference. We added weak supervision in the form of image-level tag information by setting $\tau_{e}(t,k)=(1,0)$ for the target images that do not contain the region of interest. Note that we used exactly the same class-ratio priors and weak supervision for the method in \cite{zhang2019curriculum}, for a fair comparison. Note, also, that we adopted and improved significantly the performance of the adversarial method in \cite{tsai2018learning} by using the same weak supervision information based on image-level tags, for a fair comparison\footnote{For the model in \cite{tsai2018learning}, pairs of source and target images were not used if neither had the region of interest as this confuses adversarial training, reducing its performance.}.  

\subsubsection{\textbf{Training and implementation details.}}

For all methods, the segmentation network employed was ENet \cite{paszke2016enet}, trained with the Adam optimizer \cite{Adam}, a batch size of 12 for 100 epochs, and an initial learning rate of $1\times10^{-3}$. For all adaptation models, a model trained on the source data with Eq\eqref{eq:source-sup} for 100 epochs was used as initialization. The $\lambda$ parameter in Eq\eqref{eq:ent-domain-adaption} was set empirically to $1\times10^{-2}$. For $AdaSource$, the batches used were non-aligned random slices in each domain. To learn the class-ratio prior, a ResNeXt101 \cite{resnext} regression network is used, optimized with SGD, a learning rate of $5\times 10^{-6}$, and a momentum of $0.9$. 


\subsubsection{\textbf{Evaluation.}}
The Dice similarity coefficient (DSC) and the Hausdorff distance (HD) were used as evaluation metrics in our experiments.

\subsection{Discussion} 

\begin{figure}[ht]
  \centering
  \includegraphics[width=.496\linewidth]{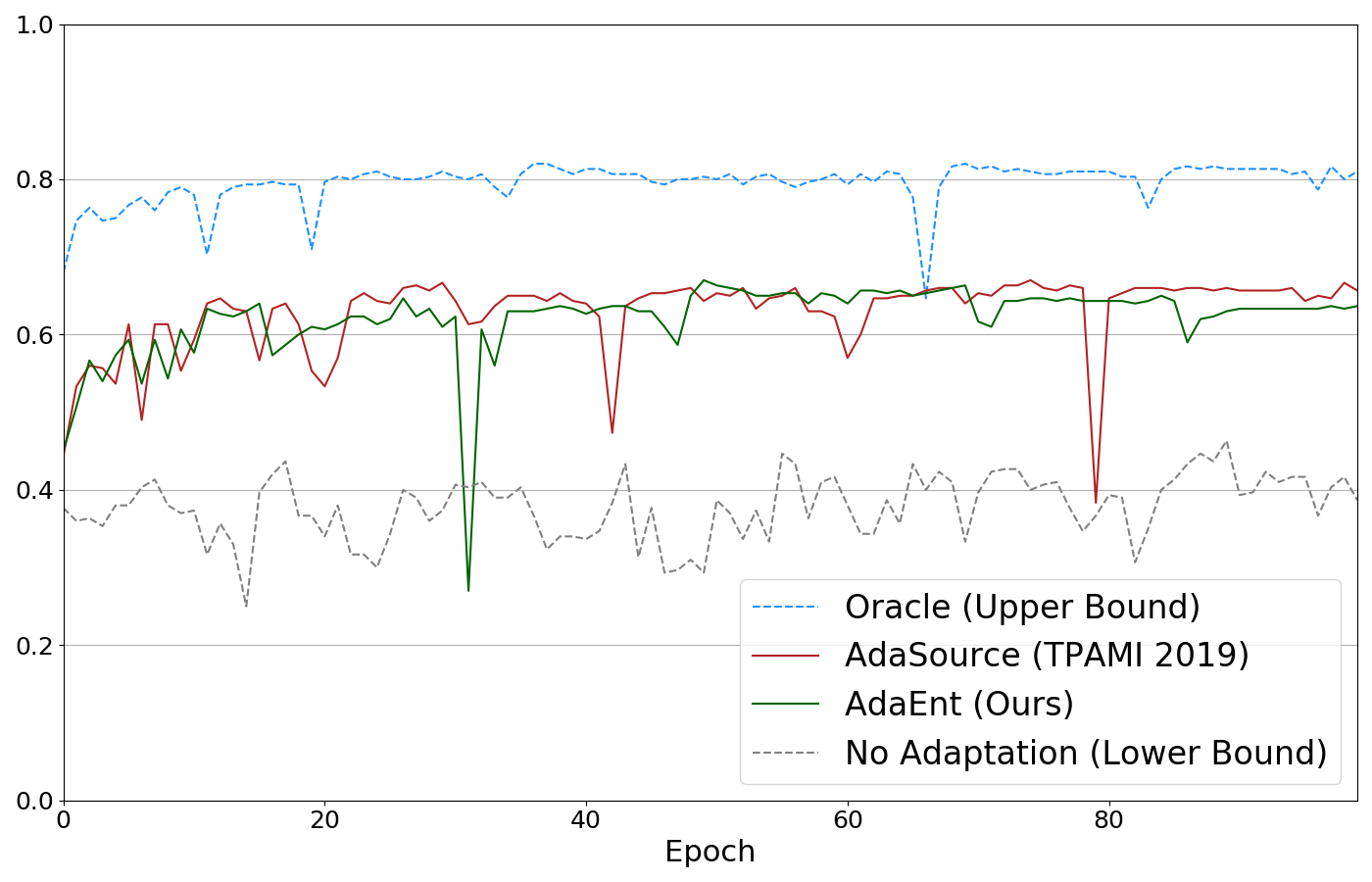}  
  \centering
  \includegraphics[width=.496\linewidth]{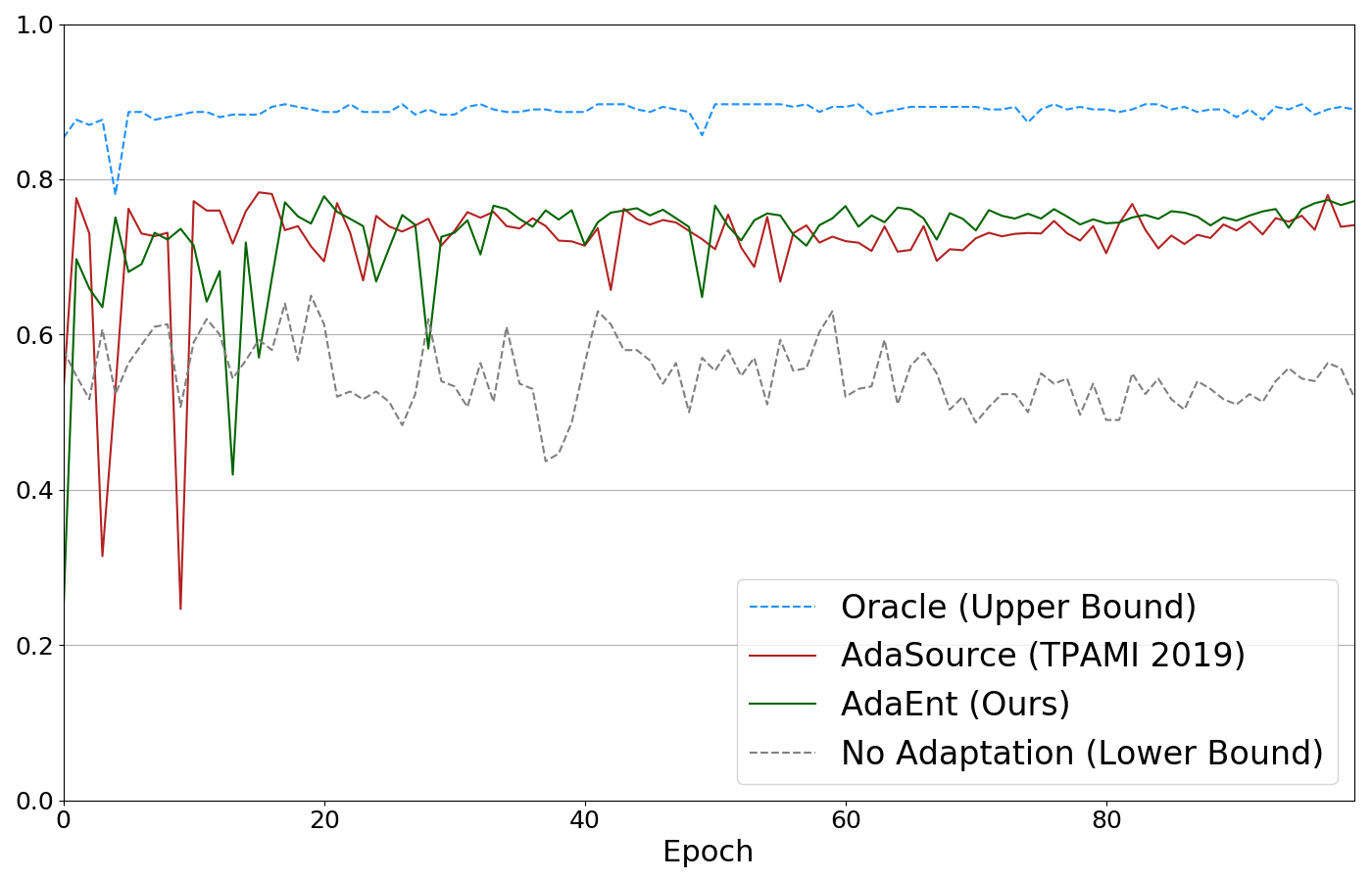}  
\caption[]{Evolution of validation DSC over training for the different models. Comparison of the proposed model to the lower and upper bounds, and to the adaptation with access to source in \cite{zhang2019curriculum} is shown. $\text{Water}\rightarrow \text{In-Phase } \textit{(left)}, \text{In-Phase} \rightarrow \text{Water } \textit{(right)}$}
\label{fig:dicefig}
\end{figure}
\begin{figure}[h!]

    \includegraphics[width=1\linewidth]{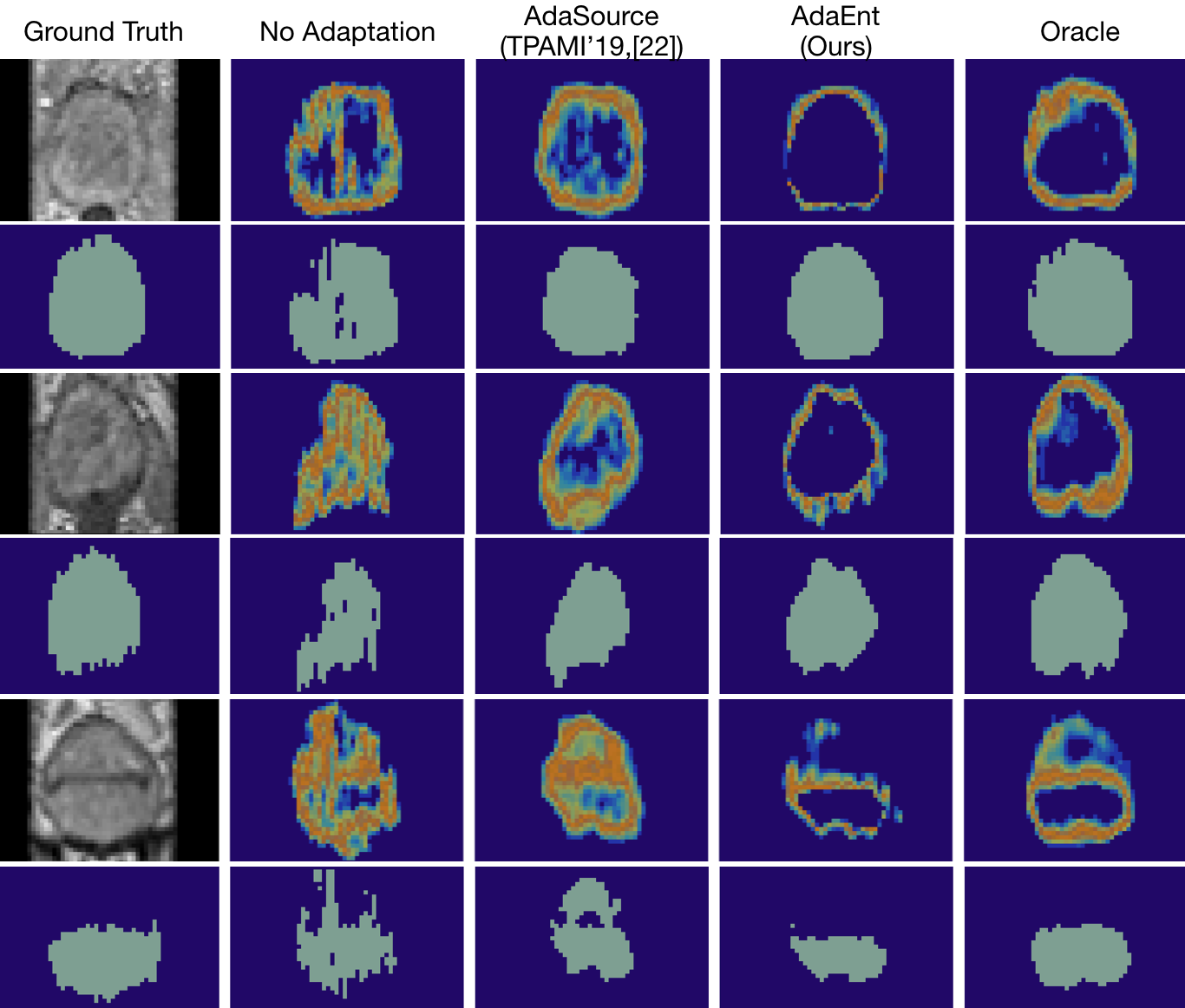}
      \caption[]{Visual results for each subject in the validation set for the several models ($\text{Water }\rightarrow \text{In-Phase} )$. First column shows an input slice and the corresponding semantic segmentation ground-truth. The other columns show segmentation results (bottom) along with prediction entropy maps produced by the different models (top). }
         \label{fig:seg}
\end{figure}

Table \ref{tab:results_doubletab} reports quantitative metrics. As expected, the model \textit{NoAdaptation}, which doesn't use any adaptation strategy but instead is trained using Eq\eqref{eq:source-sup} on the source modality, can't perform well on a different target modality. The mean DSC reached is of 46.7\% on IP, and 63.7\% on Wat, and a very high standard deviation is observed in both case, showing a high subject variability. This is confirmed in Fig. \ref{fig:seg}, where it can be seen that the output segmentation is poor on 2 out of 3 subjects. Moreover, as can be observed in Fig. \ref{fig:dicefig}, where the evolution of the training in terms of validation DSC is shown, the DSC is very unstable on the target domain throughout learning. We also observe that the adaptation task isn't symmetrically difficult, as the performance drop is much bigger in one direction (Wat $\rightarrow$ IP). The performance of  $Oracle$, the upper baseline is also lower in IP. As the visualisation in Fig. \ref{fig:s_t_im} show, the higher contrast in Wat images makes the segmentation task easier.

All models using adaptation techniques yield substantial improvement over the lower baseline. First, $Adversarial$ achieves a mean DSC of 65.3\% on IP and 77.3\% on Wat. Nonetheless, the models without adversarial strategies yield better results: $AdaSource$ achieves a mean DSC of 67.0\% on IP and 78.3\% on Wat. Interestingly, our model $AdaEnt$ shows comparable performance, with a mean DSC of 67.0\% on IP and 77.8\% on Wat. These results show that having access to more information (i.e., source data) doesn't necessarily help for the adaptation task. For both models $AdaSource$ and $AdaEnt$, the DSC comes close to the $Oracle$'s, the upper baseline, reaching respectively $82\%$ and $82\%$ of its performance respectively on IP, and $87\%$ and $89\%$ of its performance respectively on Wat. This demonstrates the efficiency of the using a class-ratio prior matching with a KL divergence. Moreover, in Fig. \ref{fig:dicefig}, we can observe that both these adaptation methods yield rapidly high validation DSC measures (first 20 epochs). This suggests that integrating such a KL divergence helps the learning process in domain adaptation.
Finally, the HD values confirm the trend across the different models. Improvement over the lower baseline model (2.45 pixels on IP, 1.44 on Wat), is substantial for both $AdaSource$ (1.34 pixels on IP, 1.14 pixels on Wat), as well as for $AdaEnt$ (1.33 pixels on IP, 1.17 on Wat). 

\begin{table}[t]
    \begin{minipage}{.5\linewidth}
\begin{tabular}{lll}
\multicolumn{3}{c}{$\text{Wat (Source) }\rightarrow \text{IP (Target) }$} \\
\midrule
Method & DSC (\%) & HD (pix) \\
\midrule
No Adaptation                                                  &  46.7 $\pm$ 10.8   & 2.45 $\pm$ 0.16   \\
Adversarial\cite{tsai2018learning} &   65.3 $\pm$ 5.5  & 1.67 $\pm$ 1.64  \\
\begin{tabular}[c]{@{}l@{}}AdaSource \cite{zhang2019curriculum}\end{tabular}    &  67.0 $\pm$ 7.2   &  1.34 $\pm$ 0.15  \\
\begin{tabular}[c]{@{}l@{}}AdaEnt (Ours)\end{tabular} &  \textbf{67.0 $\pm$ 6.1}   &  \textbf{1.33 $\pm$ 0.17} \\
Oracle &   82.3 $\pm$ 1.2  & 1.09 $\pm$ 0.16  \\
\midrule
\end{tabular}

    \end{minipage}%
    \begin{minipage}{.5\linewidth}
    \raggedleft
\begin{tabular}{lll}
\multicolumn{3}{c}{$\text{IP (Source) }\rightarrow \text{Wat (Target) }$} \\

\midrule
Method & DSC (\%) & HD (pix)\\
\midrule
No Adaptation                                                  &     63.7$\pm$ 9.1 &  1.44 $\pm$ 0.2  \\
Adversarial\cite{tsai2018learning} &   77.3 $\pm$ 7.6  & 1.15 $\pm$ 0.2   \\
\begin{tabular}[c]{@{}l@{}}AdaSource \cite{zhang2019curriculum}\end{tabular}    & \textbf{78.3 $\pm$  3.5 } & \textbf{1.14  $\pm$ 0.1} \\
\begin{tabular}[c]{@{}l@{}}AdaEnt (Ours)\end{tabular} &  77.8  $\pm$ 2.2 &  1.17 $\pm$ 0.1  \\
Oracle &   89.0 $\pm$ 2.7 & 0.90 $\pm$ 0.1  \\
\midrule
\end{tabular}
    \end{minipage} 
    \label{tab:results_doubletab}
    \caption[]{Quantitative comparisons of performance on the \textit{target} domain for the different models (mean $\pm$ std) show the efficiency of our source-relaxed formulation.}
\end{table}

Qualitative segmentations and corresponding prediction entropy maps are depicted in Figure \ref{fig:seg}, from the easiest to the hardest subject in the validation set. Without adaptation, a model trained on source data only can't recover the structure of the IVD on the target data, and is very uncertain, as revealed by the high activations in the prediction entropy maps. 
The output segmentation masks are noisy, with very irregular edges. As expected, the segmentation masks obtained using both adaptation formulations are much closer to the ground truth one, and have much more regular edges. Nonetheless, the entropy maps produced from $AdaSource$ predictions still show high entropy activations inside and close to the IVD structures. On the contrary, those produced from $AdaEnt$ look like edge detection results with high entropy activations only present along the IVD borders. Interestingly, it can be seen that even the $Oracle$'s segmentation predictions are more uncertain. This isn't surprising, as $AdaEnt$ is the only model trained to directly minimize the entropy of the predictions. The visual results confirm $AdaEnt$'s remarkable ability to produce accurate predictions with high confidence. 

\section{Conclusion}

In this paper, we proposed a simple formulation for domain adaptation (DA), which removes the need for a concurrent access to the source and target data, in the context of semantic segmentation for multi-modal magnetic resonance images. Our approach substitutes the standard supervised loss in the source domain by a direct minimization of the entropy of predictions in the target domain. To prevent trivial solutions, we integrate the entropy loss with a class-ratio prior, which is built from an auxiliary network. Unlike the recent domain-adaptation techniques, our method tackles DA without resorting to source data during the adaptation phase. Interestingly, our formulation achieved better performances than related state-of-the-art methods with access to both source and target data. 
This shows the effectiveness of our prior-aware entropy minimization and that, in several cases of interest where the domain shift is not too large, adaptation might not need access to the source data. 
Future work will address a possible class-label ratio prior shift between modalities, e.g. in case of a different field of view. Our proposed adaptation framework is usable with any segmentation network architecture.

 \section*{Acknowledgment}

This work is supported by the Natural Sciences and Engineering Research Council of Canada (NSERC), Discovery Grant program, by the The Fonds de recherche du Québec - Nature et technologies (FRQNT) grant, the Canada Research Chair on Shape Analysis in Medical Imaging, the ETS Research Chair on Artificial Intelligence in Medical Imaging, and NVIDIA with a donation of a GPU. The authors would like to thank the MICCAI 2018 IVDM3Seg organizers for providing the data.

\bibliographystyle{splncs04}
\bibliography{biblio}


\end{document}